\documentclass[sigconf]{acmart}
\usepackage[inkscapelatex=false]{svg}
\usepackage{subcaption}

\usepackage{enumitem}
\setlist{nosep}

\AtBeginDocument{%
  }

\copyrightyear{2025}
\acmYear{2025}
\setcopyright{acmlicensed}
\acmConference[MM '25] {Proceedings of the 33rd ACM International Conference on Multimedia}{October 27--31, 2025}{Dublin, Ireland.}
\acmBooktitle{Proceedings of the 33rd ACM International Conference on Multimedia (MM '25), October 27--31, 2025, Dublin, Ireland}
\acmISBN{979-8-4007-2035-2/2025/10}
\acmDOI{10.1145/3746027.3758264}

\begin{document}


\title{OpenEvents V1: Large-Scale Benchmark Dataset for Multimodal Event Grounding}


\author{Hieu Nguyen}
\orcid{0009-0001-7726-5301}
\affiliation{%
  \institution{University of Science,}
  \city{VNU-HCM}
  \country{Vietnam}
}
\author{Phuc-Tan Nguyen}
\orcid{0009-0004-0720-4038}
\affiliation{%
  \institution{University of Science,}
  \city{VNU-HCM}
  \country{Vietnam}
}
\email{}
\author{Thien-Phuc Tran}
\orcid{0009-0008-0800-6884}
\affiliation{%
  \institution{University of Science,}
  \city{VNU-HCM}
  \country{Vietnam}
}
\email{}
\author{Minh-Quang Nguyen}
\orcid{0009-0008-3520-4624}
\affiliation{%
  \institution{University of Science,}
  \city{VNU-HCM}
  \country{Vietnam}
}
\email{}
\author{Tam V. Nguyen}
\orcid{0000-0003-0236-7992}
\affiliation{%
  \institution{University of Dayton,}
  \state{Ohio}
  \country{US}
}
\email{}
\author{Minh-Triet Tran}
\orcid{0000-0003-3046-3041}
\affiliation{%
  \institution{University of Science,}
  \city{VNU-HCM}
  \country{Vietnam}
}
\email{}
\author{Trung-Nghia Le}
\orcid{0000-0002-7363-2610}
\affiliation{%
  \institution{University of Science,}
  \city{VNU-HCM}
  \country{Vietnam}
}
\email{}
\authornote{Corresponding author. Email: ltnghia@fit.hcmus.edu.vn}

\renewcommand{\shortauthors}{Hieu Nguyen et al.}

\begin{abstract}
We introduce OpenEvents V1, a large-scale benchmark dataset designed to advance event-centric vision–language understanding. Unlike conventional image captioning and retrieval datasets that focus on surface-level descriptions, OpenEvents V1 dataset emphasizes contextual and temporal grounding through three primary tasks: (1) generating rich, event-aware image captions, (2) retrieving event-relevant news articles from image queries, and (3) retrieving event-relevant images from narrative-style textual queries. The dataset comprises over 200,000 news articles and 400,000 associated images sourced from CNN and The Guardian, spanning diverse domains and time periods. We provide extensive baseline results and standardized evaluation protocols for all tasks. OpenEvents V1 establishes a robust foundation for developing multimodal AI systems capable of deep reasoning over complex real-world events. The dataset is publicly available at \url{https://ltnghia.github.io/eventa/openevents-v1}.
\end{abstract}



\begin{teaserfigure}
    \centering
    \includegraphics[width=\textwidth]{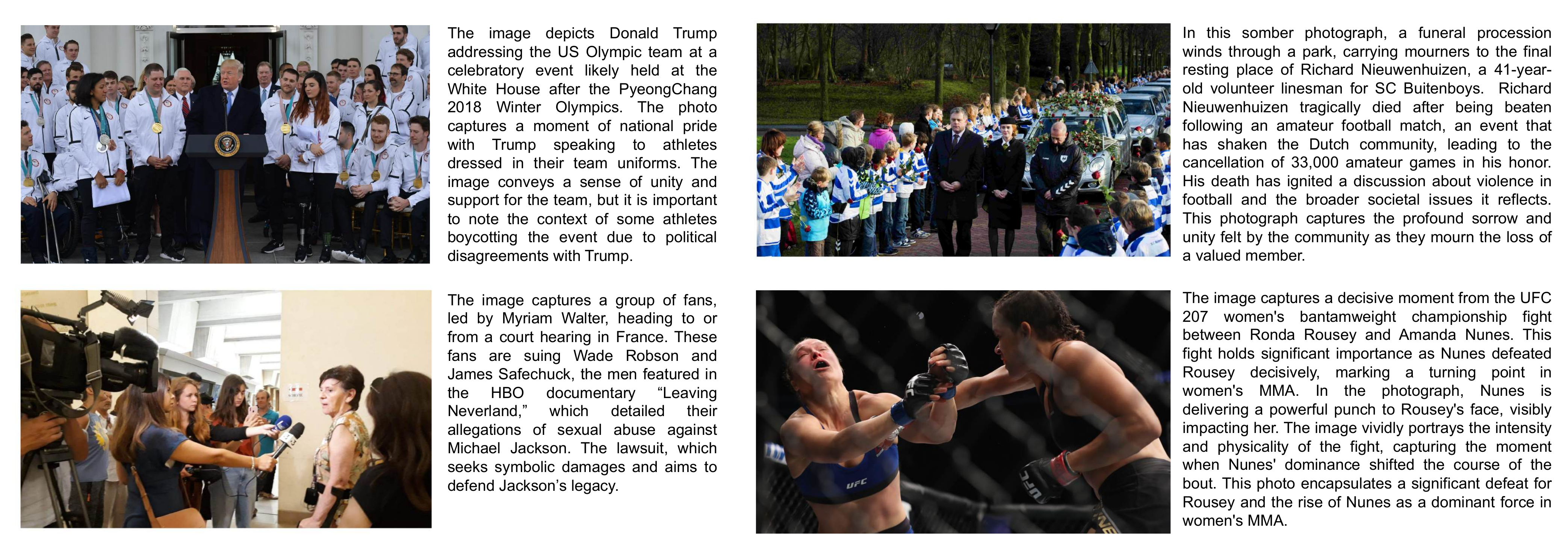}
    \caption{Sample image-caption pairs in the proposed OpenEvents V1 dataset.}
    \label{ensemblewf}
\end{teaserfigure}


\maketitle

\section{Introduction}

\begin{figure*}[t!]
    \centering
  \begin{subfigure}[t]{0.33\textwidth}
    \centering
    \includegraphics[width=\textwidth]{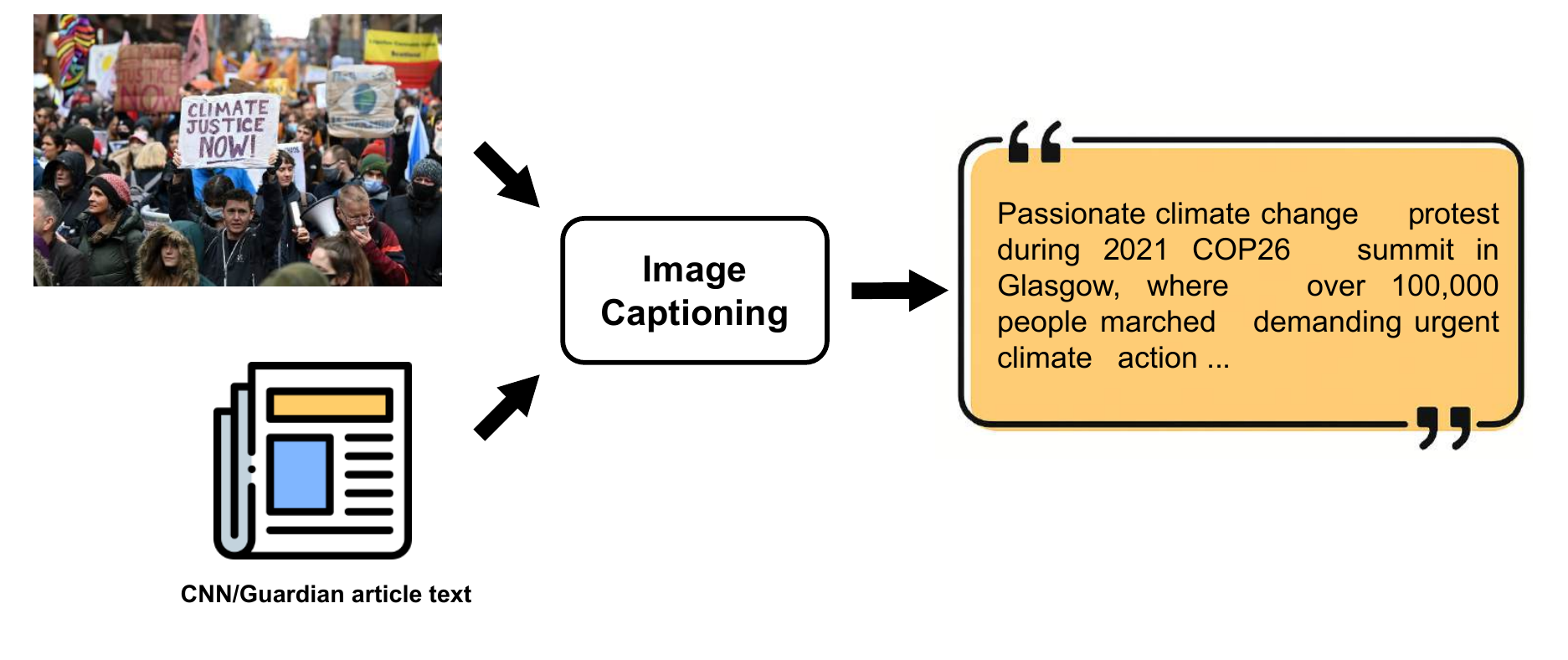}
    \caption{Event-enriched image captioning}
    \label{fig:sub_a}
  \end{subfigure}
  \hfill
  \begin{subfigure}[t]{0.33\textwidth}
    \centering
    \includegraphics[width=\textwidth]{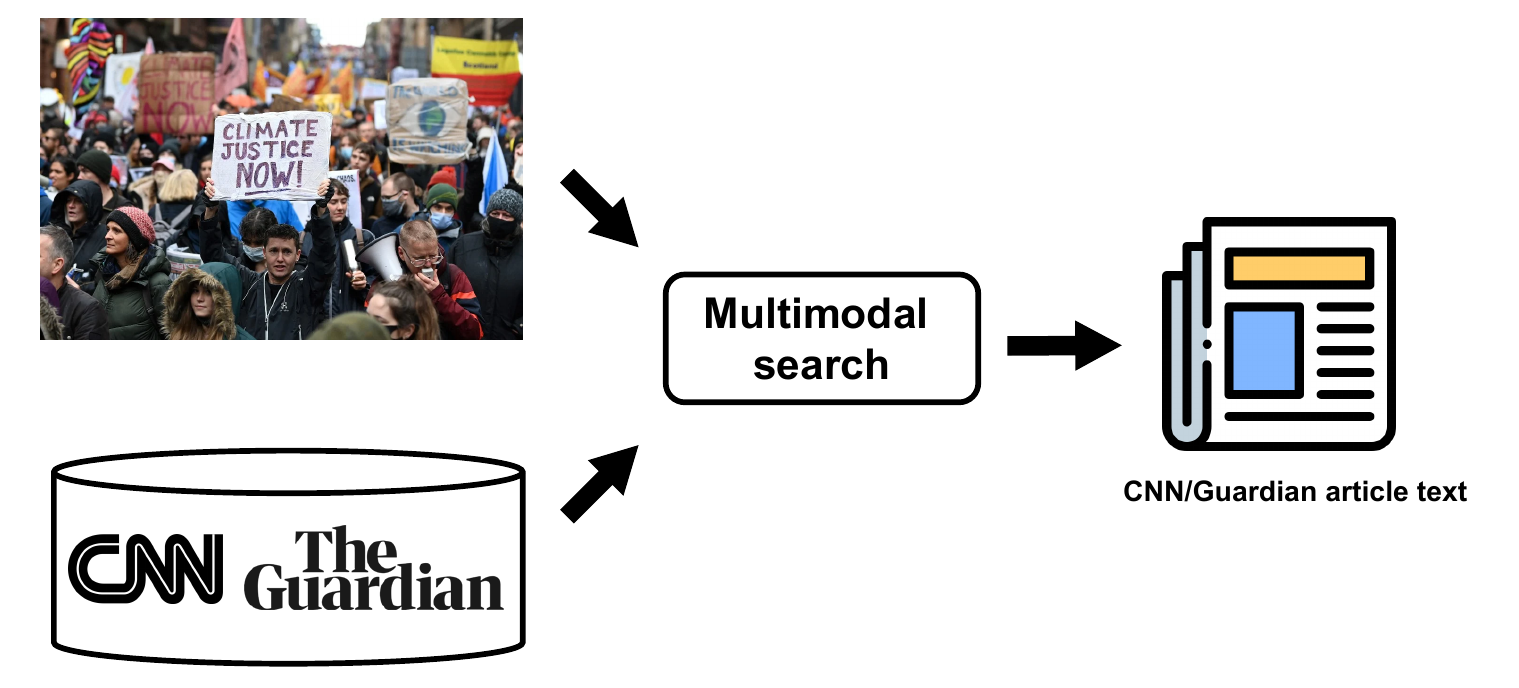}
    \caption{Event-based article retrieval}
    \label{fig:sub_b}
  \end{subfigure}
  \hfill
  \begin{subfigure}[t]{0.33\textwidth}
    \centering
    \includegraphics[width=\textwidth]{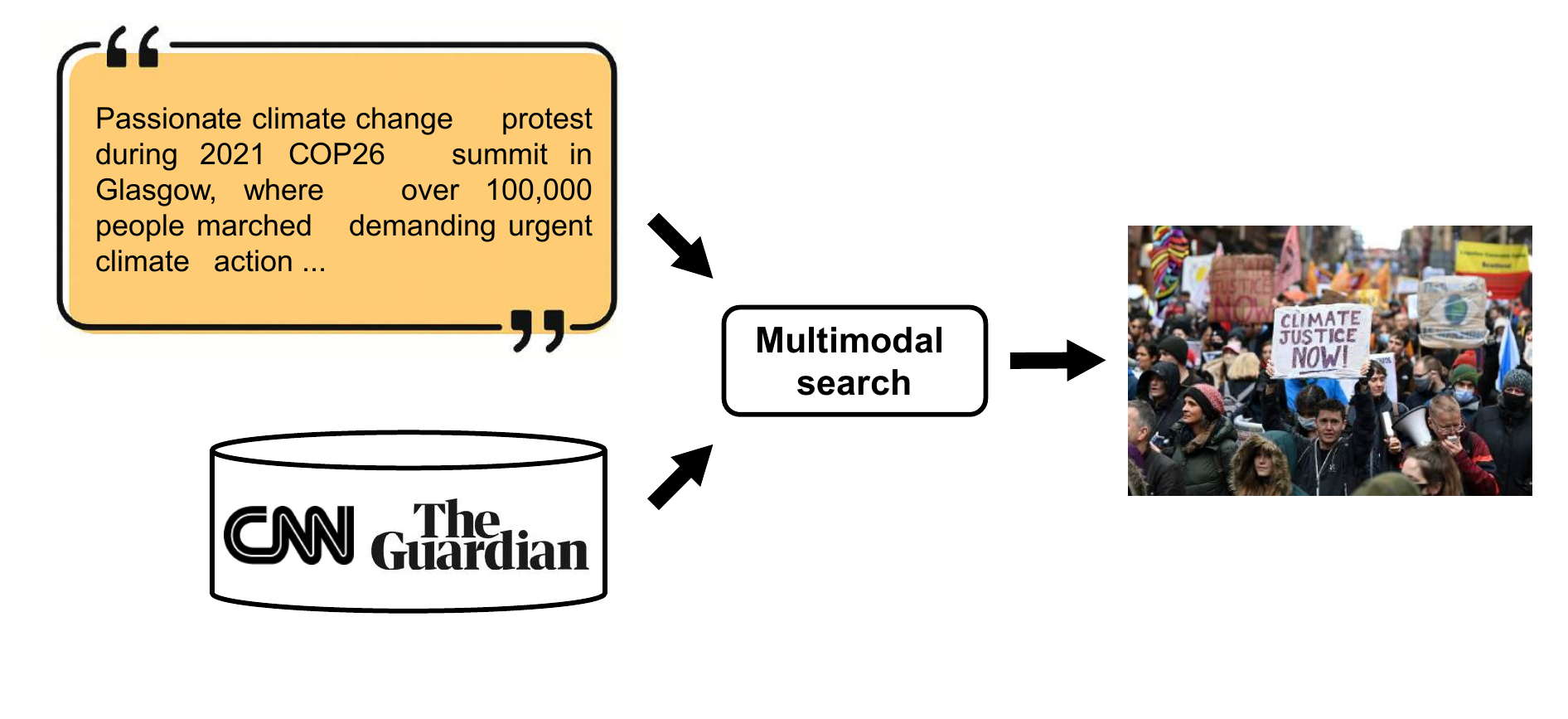}
    \caption{Event-based image retrieval}
    \label{fig:sub_c}
  \end{subfigure}
  \caption{OpenEvents V1 dataset supports three tasks for event grounding.}
  \Description{}
  \label{fig:teaser}
\end{figure*}

Recent advances in vision-language models have significantly improved the ability of machines to describe and understand visual content through natural language. However, most existing image captioning benchmarks are grounded in synthetic or simplified scenarios, focusing primarily on object-level descriptions or everyday scenes \cite{li2019coco}. These datasets and tasks fail to capture the complexity and contextual richness of real-world events, where understanding requires not just visual perception, but also temporal, social, and semantic context.

Understanding real-world events from images often requires more than just identifying objects or scenes, as it demands reasoning about what happened, who was involved, where and when it occurred, and why it matters. Such event understanding is critical in domains like journalism, historical archiving, disaster analysis, and media monitoring. However, current benchmarks lack the contextual depth needed to support these capabilities, as they rarely incorporate external knowledge or temporal dynamics. Moreover, captions generated in traditional settings are typically surface-level and visually grounded, offering limited insight into the broader narrative or consequences of the depicted moment.

To bridge this gap, we introduce the OpenEvents V1 dataset and establish a challenging, context-rich benchmark that mirrors how images are presented and interpreted in real-world settings, particularly within narrative news articles. By requiring models to jointly process both visual and textual content, this benchmark encourages the development of systems with deeper semantic grounding and event-level understanding. It further enables advancements in retrieval-augmented generation and multimodal reasoning.

\textit{Problem 1 – Event-enriched image captioning.} Given an input image, models are required to retrieve relevant contextual information from an external article database and generate a detailed, context-aware caption (\autoref{fig:sub_a}). This task encourages models to go beyond surface-level visual cues and incorporate real-world knowledge such as named entities, temporal context, and event outcomes. By grounding the caption in both visual content and textual evidence, the generated description reflects a more comprehensive understanding of the depicted event, supporting applications in visual journalism, archival, and automated media reporting.

\textit{Problem 2 – Event-based article retrieval.} In practical deployments, an image captioning system may be provided with only the image, lacking the corresponding article. Because \textit{event-enriched image captioning} requires both visual cues and contextual knowledge, the first step is to recover that missing context. To address this, we incorporate an article retrieval stage that locates the most relevant news article for the given image, ensuring the captioning model has access to the background information needed to produce accurate, event-aware descriptions (\autoref{fig:sub_b}).

\textit{Problem 3 – Event-based image retrieval.} Given a caption describing a real-world event, participants must retrieve the corresponding image from a large-scale database (\autoref{fig:sub_c}). Unlike traditional keyword-matching systems, this task requires semantic alignment between narrative-style text and event-centric visual representations. It emphasizes the ability of models to reason across modalities and handle diverse linguistic expressions of the same event. This capability is crucial for information retrieval in domains such as news curation, multimedia search engines, and digital content recommendation.

To support these tasks, OpenEvents V1 provides over 400,000  images and 200,000 articles collected from {CNN} and {The Guardian}, covering over a decade of diverse, high-impact events. Captions are generated using an improved version of the multi-stage VisChronos framework~\cite{nguyen2024vischronos}, which combines dense captioning, large language models, and contextual reasoning. In addition, we establish some baseline methods to encourage further research and benchmarking.

Our contributions are summarized as follows:

\begin{itemize}
    \item We introduce {OpenEvents V1}, a large-scale benchmark for event-enriched image captioning and event-based image retrieval.
    
    \item We define and evaluate two event grounding tasks under a unified setting with standard metrics and some baseline models.
    
    \item We present empirical findings that reveal key challenges in handling real-world, context-rich vision-language data.

    \item We release OpenEvents V1 dataset at \url{https://ltnghia.github.io/eventa/openevents-v1}.
\end{itemize}

\begin{figure*}[t!]
    \centering
    \includegraphics[width=0.8\textwidth]{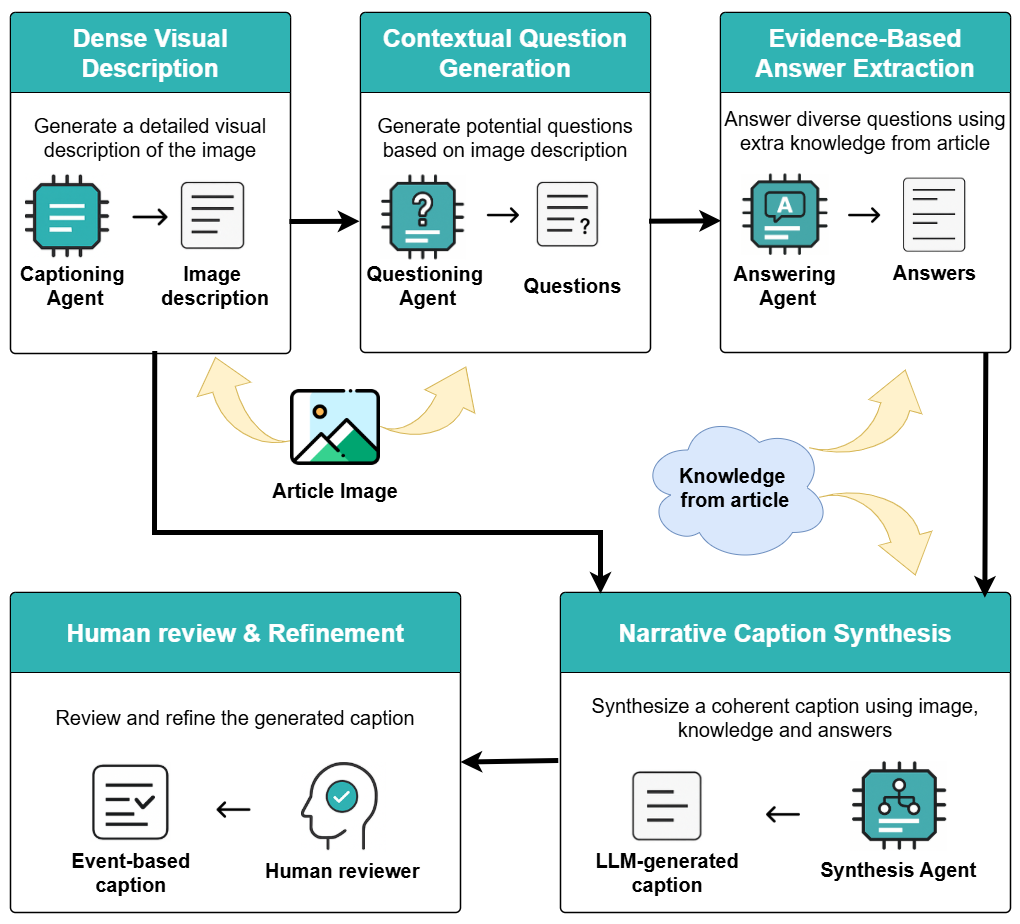}
    \caption{Human-agentic framework for dataset construction.}
    \label{fig:framework}
\end{figure*}




\section{Related Work}
\label{sec:related-work}
Image-text datasets have been foundational to the progress of multimodal learning, particularly in retrieval tasks. Early datasets such as Flickr30K \cite{lee2018stacked} provided high-quality annotated image-caption pairs, serving as standard benchmarks for evaluating cross-modal models. Building on this, large-scale datasets like WIT \cite{srinivasan2021wit} introduced multilingual and diverse image-text pairs collected from Wikipedia, enabling broader pretraining and transfer to various downstream tasks.

As the field evolved, there has been increasing interest in domain-specific datasets that serve targeted applications. For example, Yuan et al. \cite{yuan2023ramm} proposed a biomedical image-text dataset for visual question answering, while Xu et al. \cite{xu2024high} developed a dataset for agricultural disease diagnosis. These efforts demonstrate the importance of tailoring datasets to specific real-world contexts.



However, most existing datasets primarily focus on surface-level visual content, offering simple descriptions of what is visible in the image. They often lack deeper insights such as the names and attributes of entities, temporal aspects, and contextual outcomes \cite{lee2018stacked} \cite{yuan2023ramm}\cite{xu2024high}. Even large-scale datasets like WIT \cite{srinivasan2021wit} include broad coverage but provide limited depth regarding event-specific context and consequences. 
To date, no existing dataset is adequately suited to support retrieval tasks centered around real-world events, leaving a significant gap in multimodal research.


\section{OpenEvents V1 Dataset}
\label{sec:dataset}
\subsection{Dataset Construction}
\subsubsection{Data Source}
To construct a comprehensive dataset for event-based image captioning, we collected a total of 202,803 news articles and 415,324 images from two major international news outlets: {CNN} and {The Guardian}. 

\begin{figure}[t!]
    \centering
\includegraphics[width=1.0\linewidth]{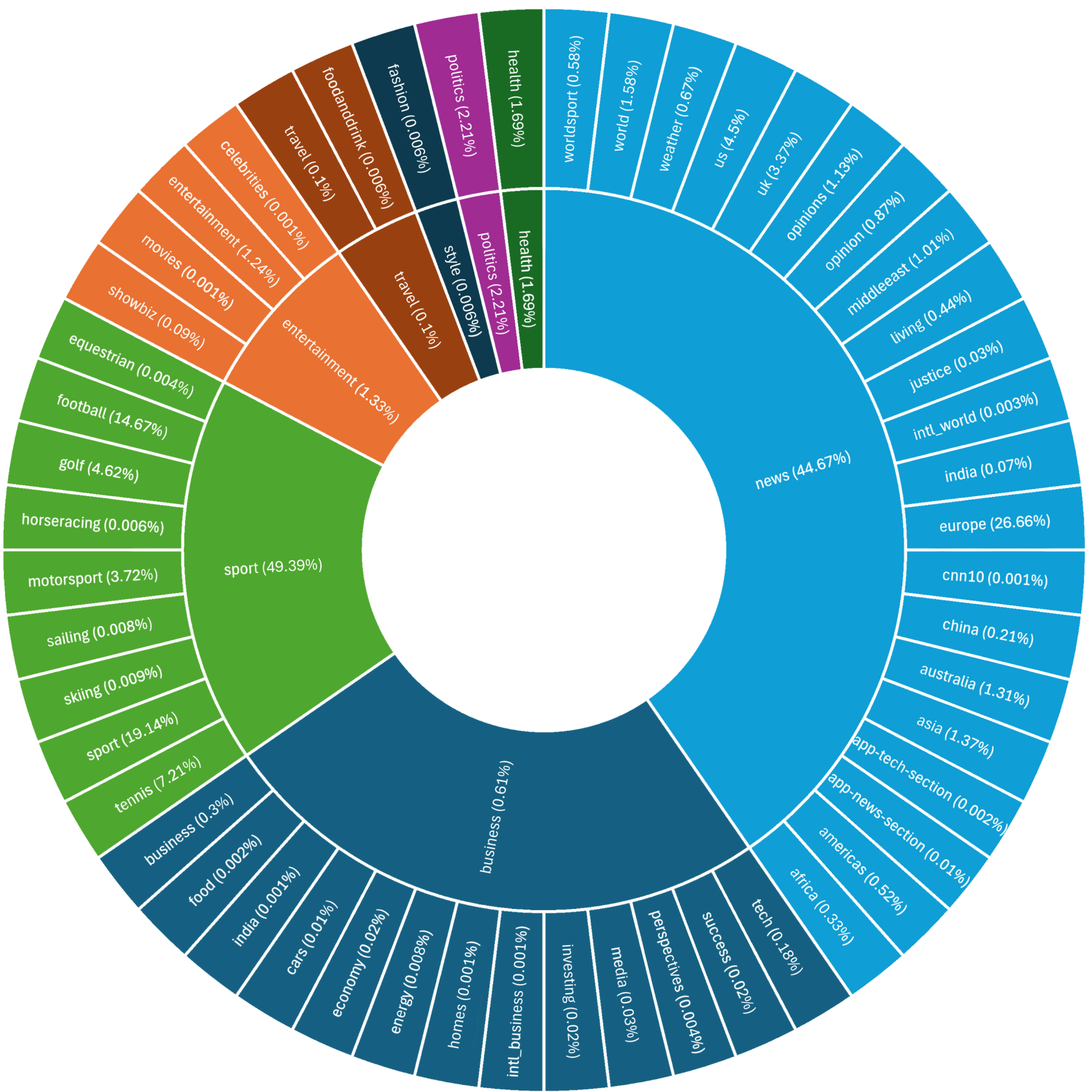}
    \caption{CNN article distribution across major content categories.}
    \label{fig:cnn-category}
\end{figure}

\begin{figure}[t!]
    \centering
\includegraphics[width=1.0\linewidth]{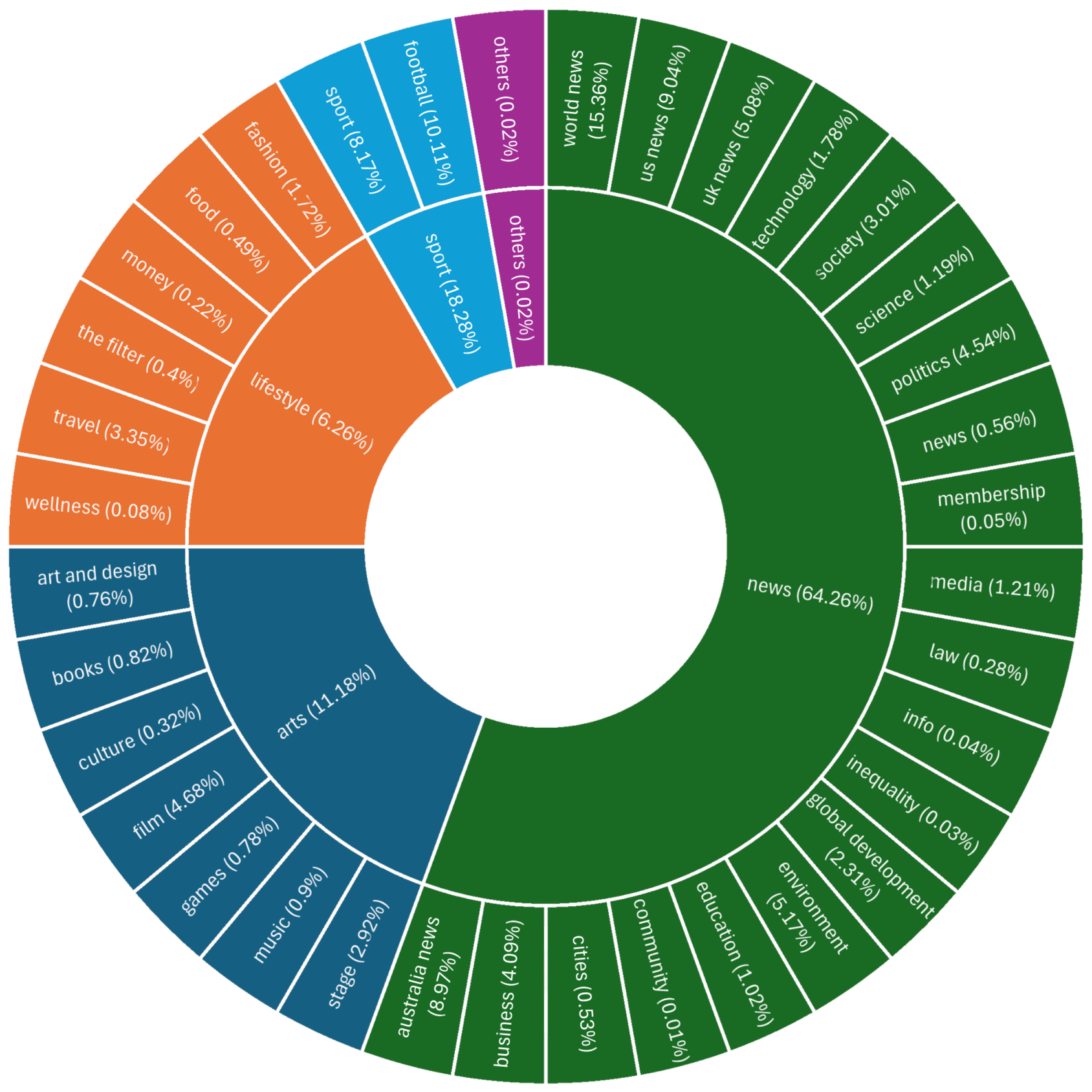}
    \caption{The Guardian article distribution across major content categories.}
    \label{fig:guardian-category}
\end{figure}

The CNN subset spans from {2011 to 2022}, while The Guardian subset covers the period from {2019 to 2025}. Both sources primarily focus on news and sports, but also contribute a rich mix of other domains such as health, politics, lifestyle, and culture as shown in Figures~\ref{fig:cnn-category} and ~\ref{fig:guardian-category}.

\subsubsection{Human-Agentic Framework}

To efficiently generate high-quality, contextually grounded captions at scale, we adopt a human-agentic framework that combines automated language models and vision-language systems with final human validation. This five-stage pipeline builds on the original VisChronos~\cite{nguyen2024vischronos}, improves into modular agents, and incorporates an explicit human-in-the-loop step for final quality control. Examples of image-caption pairs generated by our framework are illustrated in Figure~\ref{ensemblewf}.


The process consists of the following stages: (1) Dense Visual Description, (2) Contextual Question Generation, (3) Evidence-Based Answer Extraction, (4) Narrative Caption Synthesis, (5) Human Review and Refinement. This process can be seen as Figure \ref{fig:framework}

\paragraph{Step 1: Dense Visual Description}

A dense captioning model is prompted to generate a detailed visual description of the input image, capturing salient objects, people, actions, interactions, and environmental context. This output serves as the foundation for subsequent semantic reasoning.

Compared to the original VisChronos framework~\cite{nguyen2024vischronos}, which relied on proprietary paid APIs, we adopt the open-source \texttt{Molmo} \cite{deitke2024molmo} model for improved accessibility and reproducibility. Furthermore, we refine the prompting strategy to encourage more descriptive and narrative-rich outputs, rather than simple enumerations of visual elements.

\paragraph{Step 2: Contextual Question Generation}

Given the visual description, a language model generates a set of structured questions targeting the latent context behind the scene. These questions aim to uncover:
\begin{itemize}
    \item Temporal and spatial setting (when, where)
    \item Main event and causality (what, why)
    \item Participants and roles (who)
    \item Outcomes, emotions, background, and implications
\end{itemize}

This step expands the captioning task from visual perception to contextual interpretation.

\paragraph{Step 3: Evidence-Based Answer Extraction}

We provide each image with its associated article from the news corpus. The LLM is tasked with answering the generated questions by extracting sentence-level evidence from the article. Responses are only produced if the information is explicitly supported by the text. Otherwise, the model is instructed to return “no information.”

This stage ensures the factual grounding of generated content and prevents speculative generation.

\paragraph{Step 4: Narrative Caption Synthesis}

The dense description and validated answers are synthesized into a coherent, paragraph-style caption. The caption aims to narrate the event depicted in the image, incorporating relevant names, dates, locations, outcomes, and contextual signals. The result is a rich, event-enriched caption that goes far beyond conventional object-centric descriptions.




\paragraph{Step 5: Human Review and Refinement}

In the final stage, annotators assess the factual consistency with the article, correctness of named entities and temporal references, and overall readability. Captions with critical errors are edited or discarded. This procedure helps monitor quality and identify common failure patterns, contributing to the overall reliability of the dataset in practical applications such as retrieval, multimodal understanding, and media analysis.

\subsection{Dataset Description}

\begin{figure}[t!]
    \centering
\includegraphics[width=1.0\linewidth]{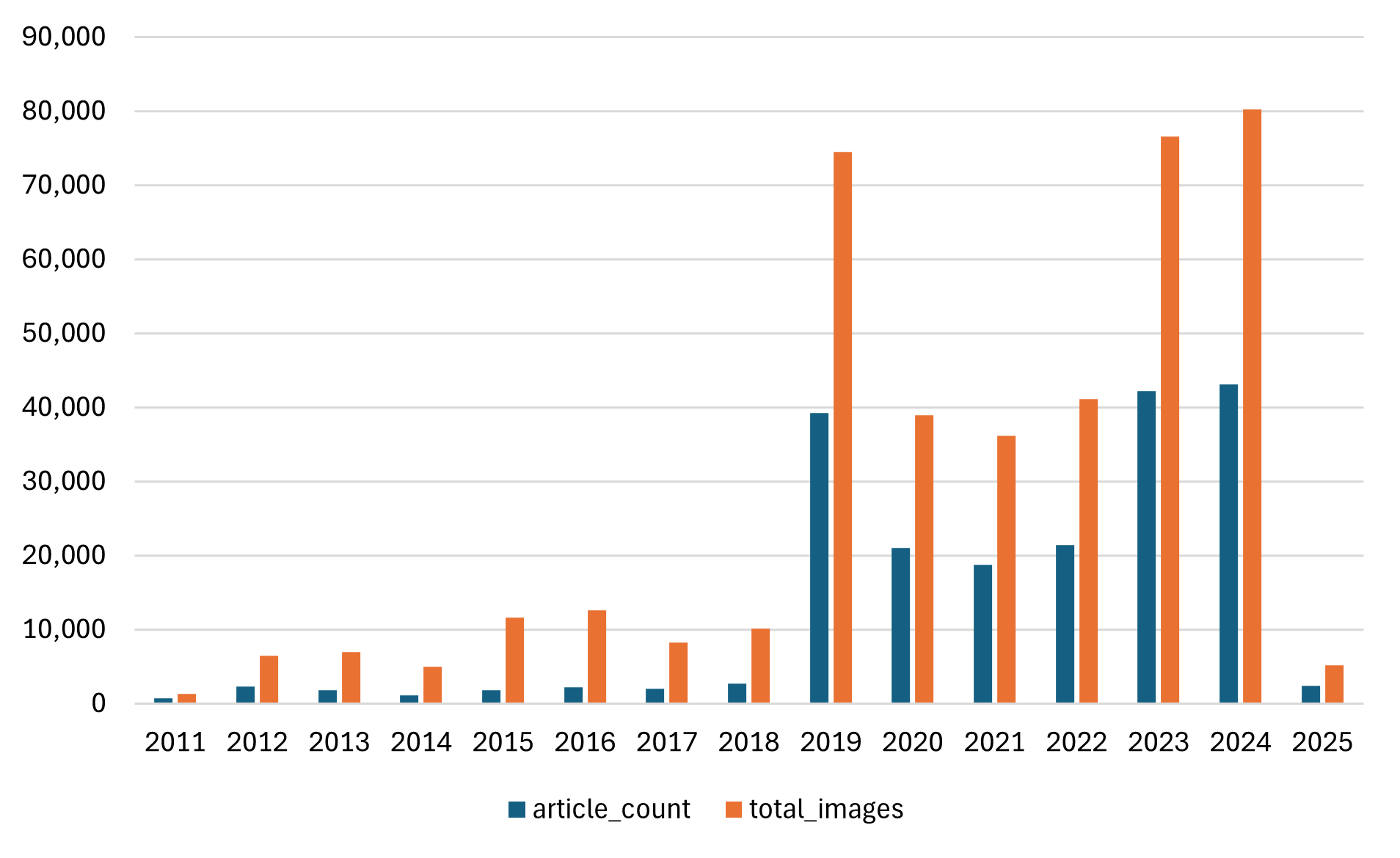}
    \caption{Distribution of articles and images by year (2011–2025).}
    \label{fig:year-cnt}
\end{figure}

\subsubsection{Statistic Information}

The CNN subset of our dataset comprises 24,200 news articles and 89,596 images, spanning from 2011 to 2022. The Guardian subset includes a much larger collection, with 178,603 articles and 325,728 images, covering 2019 to 2025.

Figure~\ref{fig:year-cnt} shows the annual distribution of articles and images. The number of samples rises sharply in 2019, then experiences a slight decline during 2020–2022, followed by a significant surge in 2023–2024, which represent the peak years in both article and image counts.

In terms of content, the dataset covers a wide range of topics. As shown in Figure~\ref{fig:cnn-category} and Figure~\ref{fig:guardian-category}, just over half of the images belong to the \texttt{news} category, followed by significant portions in \texttt{sports}. Despite this imbalance, the dataset still includes valuable long-tail sections such as \texttt{arts}, \texttt{lifestyle} \texttt{politic} and \texttt{health} offering diverse contextual grounding for caption generation.

\subsubsection{Dataset Structure}


The dataset is organized into four main subsets: \texttt{train}, \texttt{public\_test}, \texttt{private\_test}, and \texttt{database}. Each subset contains accompanying image files and structured annotations in JSON format.

\begin{itemize}
    \item \textit{Training Set}: This set contains 21,904 samples and is used for training captioning models.

    \item \textit{Public Test Set}: This set comprises 6,000 samples, this set is openly available for evaluation and can be accessed by anyone, anytime.

    \item \textit{Private Test Set}: This set consists of 4,000 samples and is reserved exclusively for the EVENTA 2025 challenge\footnote{\url{https://ltnghia.github.io/eventa/eventa-2025}}, with limited access available only during the competition period.

    \item \textit{Database}: With 415,324 images and 202,803 textual descriptions of articles, this large-scale multimodal corpus serves as the reference set for retrieval-based tasks, enabling models to search and match relevant images given a query input.
\end{itemize}

All subsets except the database share a unified image-level JSON schema consisting of two fields. The first field, \texttt{image\_id}, is a unique identifier assigned to each image. The second field, \texttt{caption}, provides the ground-truth description of the event depicted in the image.

In contrast, \texttt{database.json} operates at the article level and does not include captions directly. Instead, it provides context such as full article text and related image IDs. This structure supports a wide range of use cases, including robust training, zero-shot evaluation, and retrieval-augmented generation.

\section{Benchmarking}
\label{sec:benchmarking}

We evaluate a set of lightweight baseline models for both two problems using the public test set of OpenEvents V1. Our goal is to provide reference points that are simple, reproducible, and suitable for benchmarking future models.

Each baseline is designed with minimal resource requirements, making them accessible and reproducible for all participants. We do not fine-tune on our dataset but use pre-trained models in with several different pipelines.

\subsection{Problem 1: Event-Enriched Image Captioning}

Unlike conventional image captioning tasks that rely solely on visual content, our formulation requires generating captions grounded in both the image and its corresponding article text. This setting reflects real-world scenarios, particularly in news media, where understanding the event depicted in an image requires contextual knowledge that goes beyond what is visually observable. The model must integrate visual semantics with textual evidence such as named entities, timelines, causes, and outcomes extracted from the associated article. This joint conditioning introduces unique challenges in cross-modal reasoning, factual consistency, and contextual relevance, making the task substantially more complex than traditional vision-language generation.

We experiment with several open-source vision-language models including {SmolVLM}~\cite{marafioti2025smolvlm}, {Qwen2.5-3B}~\cite{hui2024qwen2}, and {Gemma-3-4b}~\cite{team2025gemma}, following two main pipelines: (1) \textit{Image-only}, where the model generates captions solely from the image; and (2) \textit{Image + Article}, where the model is additionally provided with the corresponding article text. The latter approach aims to simulate real-world settings where understanding an event often requires both visual perception and contextual information.

We evaluate the ability of models to generate context-rich captions that go beyond surface-level visual description by incorporating external event knowledge. To assess caption quality, we report standard captioning metrics including CLIPScore~\cite{hessel2021clipscore}, CIDEr~\cite{vedantam2015cider}, BLEU-4~\cite{papineni2002bleu} and METEOR~\cite{lavie2007meteor}, which measure fluency, descriptiveness, and semantic alignment with reference captions.

As shown in Table~\ref{tab:track1-caption}, models in the \textit{Image-only} pipeline (e.g., {SmolVLM}, {Gwen}) consistently underperform across all metrics when compared to their \textit{Image + Article} counterparts. This performance gap underscores the importance of contextual grounding. Because the ground-truth captions contain event-specific details such as named entities, temporal markers, causes, and outcomes, they cannot be accurately reconstructed from visual content alone. Incorporating article-level textual evidence is therefore crucial for generating factually grounded and semantically enriched descriptions, highlighting the value of retrieval-augmented generation in real-world image captioning scenarios.


\begin{table}[t!]
\centering
\caption{Problem 1 – Captioning results on the public test set.}
\label{tab:track1-caption}
\vspace{-3mm}
\resizebox{\columnwidth}{!}{
\begin{tabular}{lcccc}
\toprule
\textbf{Method} & \textbf{CLIPScore} & \textbf{CIDEr} & \textbf{BLEU-4} & \textbf{METEOR} \\
\midrule
SmolVLM              & 0.4609 & 0.0044  & 0.0155 & 0.0789 \\
SmolVLM + Article    & 0.5552 & 0.017   & 0.0229 & 0.0738 \\
Qwen                 & 0.5283 & 0.0282  & 0.0256 & 0.1320 \\
Qwen + Article       & 0.5855 & 0.0565  & 0.0419 & 0.1383 \\
Gemma       &0.5945        &0.0111         &0.0243        &  0.1322      \\
Gemma + Article& 0.6634  & 0.0184        & 0.0341        & 0.1453       \\

\bottomrule
\end{tabular}
}
\vspace{-3mm}
\end{table}

\subsection{Problem 2: Event-Based Article Retrieval}

\begin{table}[t!]
\centering
\caption{Problem 2 – Article retrieval results on the public test set.}
\label{tab:track1-retrieval}
\vspace{-3mm}
\small
\begin{tabular}{lcccc}
\toprule
\textbf{Method} & \textbf{mAP} & \textbf{NDCG} & \textbf{NN} & \textbf{AUC} \\
\midrule
CLIP        &0.3504 & 0.4594  &0.23  &0.0575  \\
OPEN CLIP   &0.3141  &0.4249  &0.2016  &0.0487  \\
\bottomrule
\end{tabular}
\end{table}

We evaluate CLIP\cite{radford2021learning} and OpenCLIP\cite{cherti2023reproducible} on four ranking-based metrics, such as mean Average Precision (mAP), Normalized Discounted Cumulative Gain (NDCG), Nearest Neighbor accuracy (NN), and Area Under the Precision–Recall Curve (AUC). Results in Table~\ref{tab:track1-retrieval} show a consistent advantage for CLIP across all measures, suggesting superior alignment with the event-centric properties of OpenEvents V1.

This performance difference may be attributed to differences in training data. CLIP, the model released by OpenAI, was trained on a carefully curated dataset with strong image–text correspondence. OpenCLIP, while architecturally similar, was trained on larger but less targeted public datasets, such as LAION~\cite{schuhmann2022laion}, DataComp~\cite{gadre2023datacomp}, CommonPool~\cite{ostrom2002common}, and YFCC100M~\cite{thomee2016yfcc100m}, that may not reflect the style and content of journalistic imagery. This difference in data curation likely explains CLIP’s stronger performance, reinforcing the broader insight that domain alignment in training data is critical for high retrieval accuracy in event-focused tasks.

\subsection{Problem 3: Event-Based Image Retrieval}

We evaluate the capability of models to retrieve the correct image from a large-scale visual database given a caption that narrates a real-world event. Unlike conventional retrieval tasks, where captions typically describe only the visual content, our setting presents a greater challenge: the caption is generated based on both the image and its associated article text. As a result, the descriptions often reflect deeper event semantics, such as named entities, temporal references, and contextual outcomes that may not be visually evident. This makes direct retrieval significantly harder, as models must align complex, narrative-style captions with visual content that is only partially represented in the text.

\begin{table}[t!]
\centering
\caption{Problem 3 – Image retrieval results on the public test set.}
\label{tab:track2-ranking}
\vspace{-3mm}
\resizebox{\columnwidth}{!}{
\begin{tabular}{llcccc}
\toprule
\textbf{No.} & 
\textbf{Method} & \textbf{mAP} & \textbf{NDCG} & \textbf{NN} & \textbf{AUC} \\
\midrule
1 & CLIP       & 0.2467 & 0.3407 & 0.1586 & 0.0302 \\
2 & OPEN CLIP & 0.1845 & 0.2703 & 0.1845 & 0.0185               \\
3 & SBERT + Flan T5 & 0.2134 &  0.2837 & 0.1376 & 0.0220 \\
4 & SBERT + Bart            & 0.2840 & 0.3628 & 0.1863 & 0.0372 \\
5 & SBERT + Pegasus & 0.2868 & 0.3665 & 0.1930 & 0.0362 \\
6 & SBERT + Flan T5 + CLIP         & 0.2795 & 0.3408 & 0.1986 & 0.0303 \\
7 & SBERT + Bart + CLIP & 0.3232 & 0.3978 & 0.2226 & 0.0436 \\
8 & SBERT + Pegasus + CLIP & 0.3216 & 0.3986 & 0.2173 & 0.0450 \\
\bottomrule
\end{tabular}
}
\vspace{-3mm}
\end{table}


Table~\ref{tab:track2-ranking} summarizes the results on the public test set. Article-guided retrieval consistently outperforms direct image-based approaches across all metrics, highlighting the value of contextual grounding in event-centric search.

The first two baselines (CLIP \cite{radford2021learning} and OpenCLIP \cite{cherti2023reproducible}) follow a direct approach that retrieves images purely based on query-to-image similarity. OpenCLIP \cite{cherti2023reproducible} performs worse than CLIP \cite{radford2021learning}, possibly because it was further trained on public datasets that differ from the news domain. These two methods yield relatively lower results, indicating limited capacity to align complex, event-rich queries with image content in the absence of external context.

In contrast, baselines 3 to 5 incorporate an intermediate article retrieval stage. In this approach, the system first retrieves a relevant article and then selects associated images based on their original order within that article. To facilitate this process, we employ pretrained language models such as Flan-T5, Pegasus, and BART to generate concise textual summaries for each article. During inference, the input caption is embedded using SBERT~\cite{reimers2019sentence} and matched against these summaries to identify the most semantically relevant article. This article-guided strategy yields substantial improvements in retrieval performance. Notably, the SBERT + Pegasus configuration achieves 0.2868 mAP and 0.3665 NDCG, outperforming all direct retrieval baselines.

Further improvements are observed in Baselines 6 and 7, which re-rank the top candidate images from the retrieved article using CLIP similarity. These hybrid approaches (e.g., SBERT + Bart + CLIP and SBERT + Pegasus + CLIP) achieve the highest performance across all metrics, with mAP scores exceeding 0.32 and NN accuracy above 0.22. This demonstrates the benefit of combining article-level semantic filtering with fine-grained visual similarity re-ranking.

Overall, the results highlight the importance of contextual grounding: retrieval methods that leverage both textual semantics and visual similarity consistently outperform those relying solely on direct query-to-image matching.

\section{Conclusion}
\label{sec:conclusion}

In this paper, we introduced OpenEvents V1, a large-scale benchmark designed to advance research in event-enriched image captioning and event-based image retrieval. 
To ensure high-quality outputs for event-enriched image captioning, we proposed a five-stage human–agentic framework that integrates vision–language models with article grounding and final human verification. This framework supports three benchmark tasks: event-enriched image captioning, event-based article retrieval, and event-based image retrieval. Baseline results for all tasks demonstrate the challenges and opportunities presented by this dataset, filling a critical gap in multimodal research through large-scale, real-world supervision enriched with contextual information. Our benchmark and baselines have been made publicly available to facilitate reproducibility and comparison. This dataset also has been utilized in the ACM Multimedia Grand Challenge on Event-Enriched Image Analysis (EVENTA'25)\cite{eventa25} to encourage the development of context-aware multimodal AI systems.

The OpenEvents V1 dataset can support a wide spectrum of real-world and research-oriented applications. Its dual focus on captioning and retrieval makes it an ideal testbed for evaluating retrieval-augmented generation systems and hybrid LLM architectures. The integrated article database provides a robust foundation for applications where external factual grounding is essential, enabling adaptations for visual question answering, entity extraction, and multimodal reasoning. 

Future work will focus on expanding the benchmark with multilingual content, multimodal event graphs, and video-based event understanding to capture richer temporal dynamics.

\begin{acks}
This research is supported by research funding from Faculty of Information Technology, University of Science, Vietnam National University - Ho Chi Minh City.
\end{acks}

\bibliographystyle{ACM-Reference-Format}
\balance
\bibliography{references.bib}


\begin{thebibliography}{22}


\ifx \showCODEN    \undefined \def \showCODEN     #1{\unskip}     \fi
\ifx \showISBNx    \undefined \def \showISBNx     #1{\unskip}     \fi
\ifx \showISBNxiii \undefined \def \showISBNxiii  #1{\unskip}     \fi
\ifx \showISSN     \undefined \def \showISSN      #1{\unskip}     \fi
\ifx \showLCCN     \undefined \def \showLCCN      #1{\unskip}     \fi
\ifx \shownote     \undefined \def \shownote      #1{#1}          \fi
\ifx \showarticletitle \undefined \def \showarticletitle #1{#1}   \fi
\ifx \showURL      \undefined \def \showURL       {\relax}        \fi
\providecommand\bibfield[2]{#2}
\providecommand\bibinfo[2]{#2}
\providecommand\natexlab[1]{#1}
\providecommand\showeprint[2][]{arXiv:#2}

\bibitem[Cherti et~al\mbox{.}(2023)]%
        {cherti2023reproducible}
\bibfield{author}{\bibinfo{person}{Mehdi Cherti}, \bibinfo{person}{Romain Beaumont}, \bibinfo{person}{Ross Wightman}, \bibinfo{person}{Mitchell Wortsman}, \bibinfo{person}{Gabriel Ilharco}, \bibinfo{person}{Cade Gordon}, \bibinfo{person}{Christoph Schuhmann}, \bibinfo{person}{Ludwig Schmidt}, {and} \bibinfo{person}{Jenia Jitsev}.} \bibinfo{year}{2023}\natexlab{}.
\newblock \showarticletitle{Reproducible scaling laws for contrastive language-image learning}. In \bibinfo{booktitle}{\emph{Proceedings of the IEEE/CVF conference on computer vision and pattern recognition}}. \bibinfo{pages}{2818--2829}.
\newblock


\bibitem[Deitke et~al\mbox{.}(2024)]%
        {deitke2024molmo}
\bibfield{author}{\bibinfo{person}{Matt Deitke}, \bibinfo{person}{Christopher Clark}, \bibinfo{person}{Sangho Lee}, \bibinfo{person}{Rohun Tripathi}, \bibinfo{person}{Yue Yang}, \bibinfo{person}{Jae~Sung Park}, \bibinfo{person}{Mohammadreza Salehi}, \bibinfo{person}{Niklas Muennighoff}, \bibinfo{person}{Kyle Lo}, \bibinfo{person}{Luca Soldaini}, {et~al\mbox{.}}} \bibinfo{year}{2024}\natexlab{}.
\newblock \showarticletitle{Molmo and PixMo: Open Weights and Open Data for State-of-the-Art Multimodal Models}.
\newblock \bibinfo{journal}{\emph{CoRR}} (\bibinfo{year}{2024}).
\newblock


\bibitem[Gadre et~al\mbox{.}(2023)]%
        {gadre2023datacomp}
\bibfield{author}{\bibinfo{person}{Samir~Yitzhak Gadre}, \bibinfo{person}{Gabriel Ilharco}, \bibinfo{person}{Alex Fang}, \bibinfo{person}{Jonathan Hayase}, \bibinfo{person}{Georgios Smyrnis}, \bibinfo{person}{Thao Nguyen}, \bibinfo{person}{Ryan Marten}, \bibinfo{person}{Mitchell Wortsman}, \bibinfo{person}{Dhruba Ghosh}, \bibinfo{person}{Jieyu Zhang}, {et~al\mbox{.}}} \bibinfo{year}{2023}\natexlab{}.
\newblock \showarticletitle{Datacomp: In search of the next generation of multimodal datasets}.
\newblock \bibinfo{journal}{\emph{Advances in Neural Information Processing Systems}}  \bibinfo{volume}{36} (\bibinfo{year}{2023}), \bibinfo{pages}{27092--27112}.
\newblock


\bibitem[Hessel et~al\mbox{.}(2021)]%
        {hessel2021clipscore}
\bibfield{author}{\bibinfo{person}{Jack Hessel}, \bibinfo{person}{Ari Holtzman}, \bibinfo{person}{Maxwell Forbes}, \bibinfo{person}{Ronan Le~Bras}, {and} \bibinfo{person}{Yejin Choi}.} \bibinfo{year}{2021}\natexlab{}.
\newblock \showarticletitle{CLIPScore: A Reference-free Evaluation Metric for Image Captioning}. In \bibinfo{booktitle}{\emph{EMNLP (1)}}.
\newblock


\bibitem[Hui et~al\mbox{.}(2024)]%
        {hui2024qwen2}
\bibfield{author}{\bibinfo{person}{Binyuan Hui}, \bibinfo{person}{Jian Yang}, \bibinfo{person}{Zeyu Cui}, \bibinfo{person}{Jiaxi Yang}, \bibinfo{person}{Dayiheng Liu}, \bibinfo{person}{Lei Zhang}, \bibinfo{person}{Tianyu Liu}, \bibinfo{person}{Jiajun Zhang}, \bibinfo{person}{Bowen Yu}, \bibinfo{person}{Keming Lu}, {et~al\mbox{.}}} \bibinfo{year}{2024}\natexlab{}.
\newblock \showarticletitle{Qwen2. 5-coder technical report}.
\newblock \bibinfo{journal}{\emph{arXiv preprint arXiv:2409.12186}} (\bibinfo{year}{2024}).
\newblock


\bibitem[Lavie and Agarwal(2007)]%
        {lavie2007meteor}
\bibfield{author}{\bibinfo{person}{Alon Lavie} {and} \bibinfo{person}{Abhaya Agarwal}.} \bibinfo{year}{2007}\natexlab{}.
\newblock \showarticletitle{Meteor: an automatic metric for MT evaluation with high levels of correlation with human judgments}. In \bibinfo{booktitle}{\emph{Proceedings of the Second Workshop on Statistical Machine Translation}}. \bibinfo{pages}{228--231}.
\newblock


\bibitem[Lee et~al\mbox{.}(2018)]%
        {lee2018stacked}
\bibfield{author}{\bibinfo{person}{Kuang-Huei Lee}, \bibinfo{person}{Xi Chen}, \bibinfo{person}{Gang Hua}, \bibinfo{person}{Houdong Hu}, {and} \bibinfo{person}{Xiaodong He}.} \bibinfo{year}{2018}\natexlab{}.
\newblock \showarticletitle{Stacked cross attention for image-text matching}. In \bibinfo{booktitle}{\emph{Proceedings of the European conference on computer vision (ECCV)}}. \bibinfo{pages}{201--216}.
\newblock


\bibitem[Li et~al\mbox{.}(2019)]%
        {li2019coco}
\bibfield{author}{\bibinfo{person}{Xirong Li}, \bibinfo{person}{Chaoxi Xu}, \bibinfo{person}{Xiaoxu Wang}, \bibinfo{person}{Weiyu Lan}, \bibinfo{person}{Zhengxiong Jia}, \bibinfo{person}{Gang Yang}, {and} \bibinfo{person}{Jieping Xu}.} \bibinfo{year}{2019}\natexlab{}.
\newblock \showarticletitle{COCO-CN for cross-lingual image tagging, captioning, and retrieval}.
\newblock \bibinfo{journal}{\emph{IEEE Transactions on Multimedia}} \bibinfo{volume}{21}, \bibinfo{number}{9} (\bibinfo{year}{2019}), \bibinfo{pages}{2347--2360}.
\newblock


\bibitem[Marafioti et~al\mbox{.}(2025)]%
        {marafioti2025smolvlm}
\bibfield{author}{\bibinfo{person}{Andr{\'e}s Marafioti}, \bibinfo{person}{Orr Zohar}, \bibinfo{person}{Miquel Farr{\'e}}, \bibinfo{person}{Merve Noyan}, \bibinfo{person}{Elie Bakouch}, \bibinfo{person}{Pedro Cuenca}, \bibinfo{person}{Cyril Zakka}, \bibinfo{person}{Loubna~Ben Allal}, \bibinfo{person}{Anton Lozhkov}, \bibinfo{person}{Nouamane Tazi}, {et~al\mbox{.}}} \bibinfo{year}{2025}\natexlab{}.
\newblock \showarticletitle{Smolvlm: Redefining small and efficient multimodal models}.
\newblock \bibinfo{journal}{\emph{arXiv preprint arXiv:2504.05299}} (\bibinfo{year}{2025}).
\newblock


\bibitem[Nguyen et~al\mbox{.}(2024)]%
        {nguyen2024vischronos}
\bibfield{author}{\bibinfo{person}{Phuc-Tan Nguyen}, \bibinfo{person}{Hieu Nguyen}, {and} \bibinfo{person}{Trung-Nghia Le}.} \bibinfo{year}{2024}\natexlab{}.
\newblock \showarticletitle{VisChronos: Revolutionizing Image Captioning Through Real-Life Events}. In \bibinfo{booktitle}{\emph{International Symposium on Information and Communication Technology}}. Springer, \bibinfo{pages}{127--140}.
\newblock


\bibitem[Ostrom(2002)]%
        {ostrom2002common}
\bibfield{author}{\bibinfo{person}{Elinor Ostrom}.} \bibinfo{year}{2002}\natexlab{}.
\newblock \showarticletitle{Common-pool resources and institutions: Toward a revised theory}.
\newblock \bibinfo{journal}{\emph{Handbook of agricultural economics}}  \bibinfo{volume}{2} (\bibinfo{year}{2002}), \bibinfo{pages}{1315--1339}.
\newblock


\bibitem[Papineni et~al\mbox{.}(2002)]%
        {papineni2002bleu}
\bibfield{author}{\bibinfo{person}{Kishore Papineni}, \bibinfo{person}{Salim Roukos}, \bibinfo{person}{Todd Ward}, {and} \bibinfo{person}{Wei-Jing Zhu}.} \bibinfo{year}{2002}\natexlab{}.
\newblock \showarticletitle{Bleu: a method for automatic evaluation of machine translation}. In \bibinfo{booktitle}{\emph{Proceedings of the 40th annual meeting of the Association for Computational Linguistics}}. \bibinfo{pages}{311--318}.
\newblock


\bibitem[Radford et~al\mbox{.}(2021)]%
        {radford2021learning}
\bibfield{author}{\bibinfo{person}{Alec Radford}, \bibinfo{person}{Jong~Wook Kim}, \bibinfo{person}{Chris Hallacy}, \bibinfo{person}{Aditya Ramesh}, \bibinfo{person}{Gabriel Goh}, \bibinfo{person}{Sandhini Agarwal}, \bibinfo{person}{Girish Sastry}, \bibinfo{person}{Amanda Askell}, \bibinfo{person}{Pamela Mishkin}, \bibinfo{person}{Jack Clark}, {et~al\mbox{.}}} \bibinfo{year}{2021}\natexlab{}.
\newblock \showarticletitle{Learning transferable visual models from natural language supervision}. In \bibinfo{booktitle}{\emph{International conference on machine learning}}. PmLR, \bibinfo{pages}{8748--8763}.
\newblock


\bibitem[Reimers and Gurevych(2019)]%
        {reimers2019sentence}
\bibfield{author}{\bibinfo{person}{Nils Reimers} {and} \bibinfo{person}{Iryna Gurevych}.} \bibinfo{year}{2019}\natexlab{}.
\newblock \showarticletitle{Sentence-BERT: Sentence Embeddings using Siamese BERT-Networks}. In \bibinfo{booktitle}{\emph{Proceedings of the 2019 Conference on Empirical Methods in Natural Language Processing and the 9th International Joint Conference on Natural Language Processing (EMNLP-IJCNLP)}}. \bibinfo{pages}{3982--3992}.
\newblock


\bibitem[Schuhmann et~al\mbox{.}(2022)]%
        {schuhmann2022laion}
\bibfield{author}{\bibinfo{person}{Christoph Schuhmann}, \bibinfo{person}{Romain Beaumont}, \bibinfo{person}{Richard Vencu}, \bibinfo{person}{Cade Gordon}, \bibinfo{person}{Ross Wightman}, \bibinfo{person}{Mehdi Cherti}, \bibinfo{person}{Theo Coombes}, \bibinfo{person}{Aarush Katta}, \bibinfo{person}{Clayton Mullis}, \bibinfo{person}{Mitchell Wortsman}, {et~al\mbox{.}}} \bibinfo{year}{2022}\natexlab{}.
\newblock \showarticletitle{Laion-5b: An open large-scale dataset for training next generation image-text models}.
\newblock \bibinfo{journal}{\emph{Advances in neural information processing systems}}  \bibinfo{volume}{35} (\bibinfo{year}{2022}), \bibinfo{pages}{25278--25294}.
\newblock


\bibitem[Srinivasan et~al\mbox{.}(2021)]%
        {srinivasan2021wit}
\bibfield{author}{\bibinfo{person}{Krishna Srinivasan}, \bibinfo{person}{Karthik Raman}, \bibinfo{person}{Jiecao Chen}, \bibinfo{person}{Michael Bendersky}, {and} \bibinfo{person}{Marc Najork}.} \bibinfo{year}{2021}\natexlab{}.
\newblock \showarticletitle{Wit: Wikipedia-based image text dataset for multimodal multilingual machine learning}. In \bibinfo{booktitle}{\emph{Proceedings of the 44th international ACM SIGIR conference on research and development in information retrieval}}. \bibinfo{pages}{2443--2449}.
\newblock


\bibitem[Team et~al\mbox{.}(2025)]%
        {team2025gemma}
\bibfield{author}{\bibinfo{person}{Gemma Team}, \bibinfo{person}{Aishwarya Kamath}, \bibinfo{person}{Johan Ferret}, \bibinfo{person}{Shreya Pathak}, \bibinfo{person}{Nino Vieillard}, \bibinfo{person}{Ramona Merhej}, \bibinfo{person}{Sarah Perrin}, \bibinfo{person}{Tatiana Matejovicova}, \bibinfo{person}{Alexandre Ram{\'e}}, \bibinfo{person}{Morgane Rivi{\`e}re}, {et~al\mbox{.}}} \bibinfo{year}{2025}\natexlab{}.
\newblock \showarticletitle{Gemma 3 technical report}.
\newblock \bibinfo{journal}{\emph{arXiv preprint arXiv:2503.19786}} (\bibinfo{year}{2025}).
\newblock


\bibitem[Thomee et~al\mbox{.}(2016)]%
        {thomee2016yfcc100m}
\bibfield{author}{\bibinfo{person}{Bart Thomee}, \bibinfo{person}{David~A Shamma}, \bibinfo{person}{Gerald Friedland}, \bibinfo{person}{Benjamin Elizalde}, \bibinfo{person}{Karl Ni}, \bibinfo{person}{Douglas Poland}, \bibinfo{person}{Damian Borth}, {and} \bibinfo{person}{Li-Jia Li}.} \bibinfo{year}{2016}\natexlab{}.
\newblock \showarticletitle{Yfcc100m: The new data in multimedia research}.
\newblock \bibinfo{journal}{\emph{Commun. ACM}} \bibinfo{volume}{59}, \bibinfo{number}{2} (\bibinfo{year}{2016}), \bibinfo{pages}{64--73}.
\newblock


\bibitem[Tran et~al\mbox{.}(2025)]%
        {eventa25}
\bibfield{author}{\bibinfo{person}{Thien-Phuc Tran}, \bibinfo{person}{Minh-Quang Nguyen}, \bibinfo{person}{Minh-Triet Tran}, \bibinfo{person}{Tam~V. Nguyen}, \bibinfo{person}{Trong-Le Do}, \bibinfo{person}{Duy-Nam Ly}, \bibinfo{person}{Viet-Tham Huynh}, \bibinfo{person}{Khanh-Duy Le}, \bibinfo{person}{Mai-Khiem Tran}, {and} \bibinfo{person}{Trung-Nghia Le}.} \bibinfo{year}{2025}\natexlab{}.
\newblock \showarticletitle{Event-Enriched Image Analysis Grand Challenge At ACM Multimedia 2025}. In \bibinfo{booktitle}{\emph{ACM International Conference on Multimedia}}.
\newblock


\bibitem[Vedantam et~al\mbox{.}(2015)]%
        {vedantam2015cider}
\bibfield{author}{\bibinfo{person}{Ramakrishna Vedantam}, \bibinfo{person}{C Lawrence~Zitnick}, {and} \bibinfo{person}{Devi Parikh}.} \bibinfo{year}{2015}\natexlab{}.
\newblock \showarticletitle{Cider: Consensus-based image description evaluation}. In \bibinfo{booktitle}{\emph{Proceedings of the IEEE conference on computer vision and pattern recognition}}. \bibinfo{pages}{4566--4575}.
\newblock


\bibitem[Xu et~al\mbox{.}(2024)]%
        {xu2024high}
\bibfield{author}{\bibinfo{person}{Jiaxin Xu}, \bibinfo{person}{Hongliang Zhou}, \bibinfo{person}{Yufan Hu}, \bibinfo{person}{Yongfei Xue}, \bibinfo{person}{Guoxiong Zhou}, \bibinfo{person}{Liujun Li}, \bibinfo{person}{Weisi Dai}, {and} \bibinfo{person}{Jinyang Li}.} \bibinfo{year}{2024}\natexlab{}.
\newblock \showarticletitle{High-Accuracy Tomato Leaf Disease Image-Text Retrieval Method Utilizing LAFANet}.
\newblock \bibinfo{journal}{\emph{Plants}} \bibinfo{volume}{13}, \bibinfo{number}{9} (\bibinfo{year}{2024}), \bibinfo{pages}{1176}.
\newblock


\bibitem[Yuan et~al\mbox{.}(2023)]%
        {yuan2023ramm}
\bibfield{author}{\bibinfo{person}{Zheng Yuan}, \bibinfo{person}{Qiao Jin}, \bibinfo{person}{Chuanqi Tan}, \bibinfo{person}{Zhengyun Zhao}, \bibinfo{person}{Hongyi Yuan}, \bibinfo{person}{Fei Huang}, {and} \bibinfo{person}{Songfang Huang}.} \bibinfo{year}{2023}\natexlab{}.
\newblock \showarticletitle{Ramm: Retrieval-augmented biomedical visual question answering with multi-modal pre-training}. In \bibinfo{booktitle}{\emph{Proceedings of the 31st ACM International Conference on Multimedia}}. \bibinfo{pages}{547--556}.
\newblock


\end{thebibliography}
\end{document}